\documentclass[a4paper,conference]{IEEEtran}
\IEEEoverridecommandlockouts
\usepackage{cite}
\usepackage{amsmath,amssymb,amsfonts}
\usepackage{algorithmic}
\usepackage{graphicx}
\usepackage{textcomp}
\usepackage{xcolor}
\usepackage{url}
\usepackage[hang,small,bf]{caption}
\usepackage[subrefformat=parens]{subcaption}
\def\BibTeX{{\rm B\kern-.05em{\sc i\kern-.025em b}\kern-.08em
    T\kern-.1667em\lower.7ex\hbox{E}\kern-.125emX}}
\begin{document}

\title{Space evaluation based on pitch control\\using drone video in Ultimate 
\thanks{This work was supported by JSPS KAKENHI 23H03282 and JST Presto JPMJPR20CA.}
}

\author{
\IEEEauthorblockN{Shunsuke IWASHITA}
\IEEEauthorblockA{\textit{Nagoya University} \\
Nagoya, Japan \\
iwashita.shunsuke@g.sp.m.is.nagoya-u.ac.jp}
\and
\IEEEauthorblockN{Atom SCOTT}
\IEEEauthorblockA{\textit{Nagoya University} \\
Nagoya, Japan \\
atom.james.scott@gmail.com}
\and
\IEEEauthorblockN{Rikuhei UMEMOTO}
\IEEEauthorblockA{\textit{Nagoya University} \\
Nagoya, Japan \\
umemoto.rikuhei@g.sp.m.is.nagoya-u.ac.jp}
\and
\IEEEauthorblockN{\makebox[\dimexpr\linewidth-2em][c]{%
\begin{tabular}{c@{\hspace{5em}}c}
Ning DING & Keisuke FUJII \\
\fontsize{10}{11}\selectfont \textit{Nagoya Institute of Technology} & \fontsize{10}{11}\selectfont \textit{Nagoya University} \\
\fontsize{10}{11}\selectfont Nagoya, Japan & \fontsize{10}{11}\selectfont Nagoya, Japan \\
\fontsize{10}{11}\selectfont ding.ning@nitech.ac.jp & \fontsize{10}{11}\selectfont fujii@i.nagoya-u.ac.jp
\end{tabular}
}}
}

\maketitle
\vspace{-5pt}
\begin{abstract} 
Ultimate is a sport in which teams of seven players compete for points by passing a disc into the end zone. A distinctive aspect of Ultimate is that the player holding the disc is unable to move, underscoring the significance of creating space to receive passes. Despite extensive research into space evaluation in sports such as football and basketball, there is a paucity of information available for Ultimate. This study focuses on the 3-on-3 format, which is widely practiced in Ultimate, and evaluates space during offensive play. The data collection process entailed the use of drones for filming and the subsequent correction of the angles for the purpose of obtaining positional data. The model is derived from the pitch control model of soccer and adapted to the rules of Ultimate, where the player holding the disc is stationary. The integration of position and distance weights with pitch control values enables the derivation of space evaluation metrics. The findings of this study indicate that movement to create space and accurate passing into that space are both significant factors in scoring.
The code is available at \url{https://github.com/shunsuke-iwashita/USO}.
\end{abstract}

\vspace{-2pt}
\section{Introduction}
\vspace{-2pt}

Ultimate is a sport in which two teams of seven players each compete for points. The game is played on a 100 $\times$ 37 m rectangular court, with 18 m on either side of the long side called the end zone. Each team scores points by catching the disc in the opponent's end zone. Since players cannot move while holding the disc, the offense proceeds only by passing and catching the disc, aiming for the opponent's end zone. Ultimate is a non-contact sport that avoids full contact. There is no taking of the disc from the opponent, and no tackling or other contact is permitted during the contested catch. Therefore, it is essential that the player has an advantageous position over his/her opponent. In addition, understanding the advantageous space is directly related to increasing scoring opportunities. In this study, we clarify the spaces in which the offense has an advantage and examine the quantitative evaluation of these spaces.

In Ultimate, there remains a lack of research on space evaluation. In contrast, many other sports have focused on player positioning and space evaluation.  The basis of dominant region research is the study of Voronoi diagrams, which divide regions according to which points are closest to each other. 
The minimum arrival time of the Voronoi diagram \cite{taki1996development}, which takes into account velocity and acceleration, has been studied since early times. 
Such studies have used kinematic models to calculate velocity and acceleration for each of the previous players, but there have also been studies that estimated the kinematic model for each player \cite{brefeld2019probabilistic}. 
More advanced studies have quantified off-ball scoring opportunities with probabilistic physics-based models \cite{spearman2018beyond}.
Following this research, some studies evaluate the optimal position of a defender \cite{umemoto2023evaluation} or a player who sacrifices himself for his teammates without receiving the ball \cite{teranishi2022evaluation}. While these studies have provided tactical insight and a better understanding of optimal space utilization in soccer and other sports, such studies have not yet progressed in Ultimate.

In this study, we adapt the Potential Pitch Control Field (PPCF) used in OBSO \cite{spearman2018beyond}, which is modeled consistently up to scoring probability, to the Ultimate rules and propose an Ultimate space evaluation index based on it. The contributions of this study are:
(1) Pioneering efforts in the study of Ultimate play analytics, (2) Adaptation of the space evaluation metric for soccer to Ultimate, and (3) Showing the result that a pass to a point with a high space evaluation value may increase the probability of scoring a goal.

\vspace{-6pt}
\section{Proposed method}
\vspace{-4pt}
\subsection{Adaptation of PPCF to Ultimate}
\begin{figure}[ht]
    \centering
    \begin{subfigure}[b]{0.45\textwidth}
        \centering
        \includegraphics[width=\textwidth]{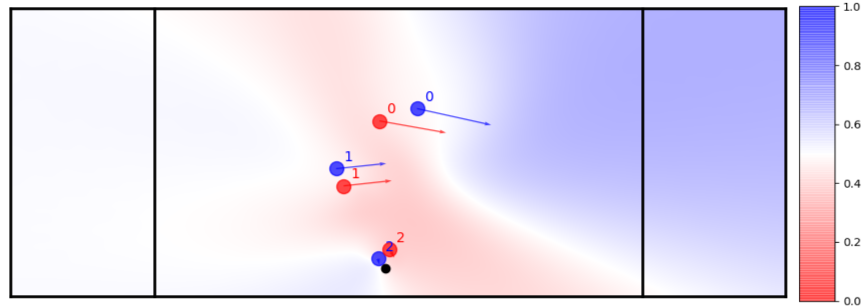}
        \caption{PPCF before modification}
        \label{fig:ppcf_a}
    \end{subfigure}
    \hfill
    \begin{subfigure}[b]{0.45\textwidth}
        \centering
        \includegraphics[width=\textwidth]{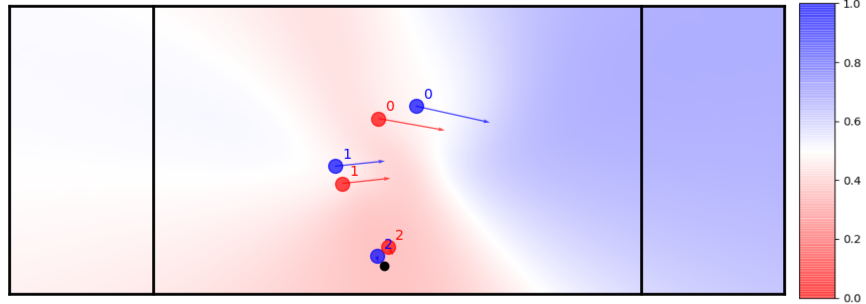}
        \caption{PPCF adopted for Ultimate}
        \label{fig:ppcf_b}
    \end{subfigure}
    \hfill
    \begin{subfigure}[b]{0.3\textwidth}
        \centering
        \includegraphics[width=\textwidth]{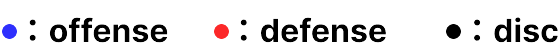}
    \end{subfigure}
    \caption{Comparison of PPCF before and after modification for Ultimate}
    \label{fig:ppcf}
\end{figure}

\vspace{-2pt}
First, the space evaluation metrics of soccer \cite{spearman2018beyond}
are adapted to the rules and characteristics of Ultimate. PPCF is the metric used in \cite{spearman2018beyond} and is the probability that a player controls a point on the field at a given time.
In soccer, the PPCF (Fig. \ref{fig:ppcf}a) is calculated for all players. However, in order to adapt to the characteristics of Ultimate, where the player holding the disc cannot move, this player is excluded from the calculation of the PPCF. Additionally, defenders within three meters of the player holding the disc were excluded from the PPCF calculation (Fig. \ref{fig:ppcf}b). This is because they are essentially performing a stalling action that limits the path of the pass.

\vspace{-20pt}
\subsection{USO (Ultimate Scoring Opportunity)}
\vspace{-2pt}
We propose the Ultimate Scoring Opportunity (USO), which incorporates both scoring opportunities and the value of the area, into the PPCF, which previously represented the value of the area alone. The USO is calculated by multiplying a position weight ($w_{area}$) and a passing distance weight ($w_{distance}$). $w_{area}$ considers the ease of scoring and passing on the field and is set to 1 inside the end zone. The remaining points are normalized to the angle of the end zone as seen from that point. $w_{distance}$ accounts for the difficulty of passing based on distance and is a normalized value of the distance from the point to the player holding the disc.

\vspace{-2pt}
\section{Experiments}
\vspace{-4pt}
\subsection{Data collection}
\vspace{-2pt}
Ultimate 3-on-3 was filmed by a drone (Mavic 3, DJI). The court was a rectangle measuring 54 m by 20 m, with 10 m at each end designated as the end zone. Following filming, the start and end times of each set of 3-on-3 were recorded and subsequently divided into 262 videos.
Automatic tracking was attempted to obtain positional data, but this proved challenging due to the limited visual information resulting from the small size of the players and disks in the video. Consequently, we manually annotated the bounding box with eight selected videos.
The eight videos consisted of four scoring sets and four turnover sets (a turnover is defined as the change from offense to defense).

To transform the positional coordinates of the players and disc in the image into those of a 3-on-3 court, a homography transformation was performed using the Sportslabkit library \cite{scott2022soccertrack}. The center of the bounding box was used as the positional coordinates, and the direction of the offense was standardized.

\vspace{-2pt}
\subsection{Evaluation of the relationship between scores and USO}
\vspace{-2pt}
First, the maximum value of USO in each frame of each set was used as the USO score. 
The last three passes in each set were defined as the last pass, second last pass, and third last pass, in order from the last. The average USO Score in frames 30-21, 20-11, and 10-1, which were the three segments of the 30 frames immediately preceding each pass, were compared between the scoring sets and the turnover sets. The distance between the point of maximum USO and the point where the pass was actually made, and the difference in USO between the two points at the time the pass was made, were also compared in the same way. The results are shown in Table \ref{tab:eval_result}. The values in columns 3-5 from the left are the average of the USO scores for the four sets, and the distance between the two points is the distance on the court. The table shows that the USO Score increased just before the last pass in the scoring set, indicating that space was created just before the last pass. In addition, the last pass was made close to the point where the USO was highest. On the other hand, there was no increase in USO Score in the turnover set, and the last pass was made relatively far from the point where the USO was highest.

\begin{table}[htb]
\centering
\caption{Average USO Score and the difference between the point of maximum USO and the point where the pass was made}
{
\small
\begin{tabular}{ccccccc} \hline
    Pass & Result & \multicolumn{3}{c}{Frame} & \scalebox{0.8}{Dist. from} & \scalebox{0.8}{USO} \\ \cline{3-5}
    & & 30-21 & 20-11 & 10-1 & \scalebox{0.8}{the max. (m)} & \scalebox{0.8}{difference} \\ \hline
    \scalebox{1}{Third} & Score & 0.400 & 0.399& 0.401 & 15.2 & 0.141 \\
    last & \scalebox{0.8}{Turnover} & 0.296 & 0.335 & 0.368 & 18.3 & 0.149 \\
    &&& \\
    \scalebox{1}{Second } & Score & 0.468 & 0.437 & 0.395 & 5.8 & 0.034 \\
    last & \scalebox{0.8}{Turnover} & 0.419 & 0.396 & 0.384 & 13.6 & 0.115 \\
    &&& \\
    Last & Score & 0.536 & 0.576 & 0.597 & 3.8 & 0.068 \\
    & \scalebox{0.8}{Turnover} & 0.521 & 0.526 & 0.518 & 8.6 & 0.124 \\ \hline
\end{tabular}
}
\label{tab:eval_result}
\end{table}

\vspace{-6pt}
\section{Conclusion}
\vspace{-2pt}
We proposed an Ultimate's space evaluation index, USO. The following factors were identified as contributing to the score: (1) movement to create space, and (2) accurate passing into space. Further application to 7-on-7 and adaptation to the unique characteristics of Ultimate are essential for future developments.


\vspace{-4pt}
\begin{thebibliography}{99}
\bibitem{taki1996development}
Taki, Tsuyoshi, Jun-ichi Hasegawa, and Teruo Fukumura. "Development of motion analysis system for quantitative evaluation of teamwork in soccer games." Proceedings of 3rd IEEE international conference on image processing. Vol. 3. IEEE, 1996.
\bibitem{brefeld2019probabilistic}
Brefeld, Ulf, Jan Lasek, and Sebastian Mair. "Probabilistic movement models and zones of control." Machine Learning 108.1 (2019): 127-147.
\bibitem{spearman2018beyond}
Spearman, William. "Beyond expected goals." Proceedings of the 12th MIT sloan sports analytics conference. 2018.
\bibitem{umemoto2023evaluation}
Umemoto, Rikuhei, and Keisuke Fujii. "Evaluation of Team Defense Positioning by Computing Counterfactuals using StatsBomb 360 data."
\bibitem{teranishi2022evaluation}
Teranishi, Masakiyo, et al. "Evaluation of creating scoring opportunities for teammates in soccer via trajectory prediction." International Workshop on Machine Learning and Data Mining for Sports Analytics. Cham: Springer Nature Switzerland, 2022.
\bibitem{scott2022soccertrack}
Scott, Atom, et al. "SoccerTrack: A dataset and tracking algorithm for soccer with fish-eye and drone videos." Proceedings of the IEEE/CVF Conference on Computer Vision and Pattern Recognition. 2022.
\end{thebibliography}
\end{document}